\documentclass[letterpaper, 10 pt, conference]{ieeeconf}  

\IEEEoverridecommandlockouts                              

\usepackage{booktabs,multirow}
\usepackage{algorithm}
\usepackage[noend]{algpseudocode}
\usepackage[english]{babel}
\usepackage[pdftex]{graphicx}
\usepackage{adjustbox}
\usepackage[colorlinks,linkcolor=red]{hyperref}
\DeclareGraphicsExtensions{.pdf,.jpg,.jpeg,.png}

\usepackage{graphicx} 
\usepackage{color, colortbl}
\usepackage{bm}
\usepackage[caption=false]{subfig}
\usepackage{amsmath} 
\usepackage{amssymb}  
\usepackage{physics}
\usepackage[table]{xcolor}
\usepackage{algcompatible}
\usepackage{multicol}
\usepackage{graphicx}
\usepackage{amsmath}
\usepackage[font=small,labelfont=bf]{caption}
\usepackage{bm}
\usepackage{glossaries}
\usepackage{float}
\usepackage[nodisplayskipstretch]{setspace}
\usepackage{epsfig} 
\usepackage{mathptmx} 
\usepackage{times} 
\newacronym{dofs}{DoFs}{Degrees-of-Freedoms}

\makeatletter
\newcounter {subsubsubsection}[subsubsection]
\renewcommand\thesubsubsubsection{\thesubsubsection .\@arabic\c@subsubsubsection}
\newcommand\subsubsubsection{\@startsection{subsubsubsection}{4}{\z@}%
	{-3.25ex\@plus -1ex \@minus -.2ex}%
	{1.5ex \@plus .2ex}%
	{\normalfont\normalsize\bfseries}}
\newcommand*\l@subsubsubsection{\@dottedtocline{3}{10.0em}{4.1em}}
\newcommand*{\subsubsubsectionmark}[1]{}
\makeatother

\newcommand{\trsp}{{\scriptscriptstyle\top}}

\newcommand{\minus}{\scalebox{0.75}[1.0]{$-$}}

\graphicspath{{./figures/}} 

\title{\LARGE \bf
	Contact Optimization with Learning from Demonstration: \\ Application in Long-term Non-prehensile Planar Manipulation
}

\author{Teng Xue and Sylvain Calinon 
\thanks{The authors are with the Robot Learning and Interaction Group, Idiap Research Institute, Martigny, Switzerland and EPFL, Lausanne, Switzerland; (e-mail: firstname.lastname@idiap.ch)} 
}

\begin{document}

\maketitle
\thispagestyle{empty}
\pagestyle{empty}

\begin{abstract}
	
Long-term non-prehensile planar manipulation is a challenging task for planning and control, requiring determination of both continuous and discrete contact configurations, such as contact points and modes. This leads to the non-convexity and hybridness of contact optimization. To overcome these difficulties, we propose a novel approach that incorporates human demonstrations into trajectory optimization. We show that our approach effectively handles the hybrid combinatorial nature of the problem, mitigates the issues with local minima present in current state-of-the-art solvers, and requires only a small number of demonstrations while delivering robust generalization performance. We validate our results in simulation and demonstrate its applicability on a pusher-slider system with a real Franka Emika robot.

\end{abstract}

\section{Introduction}
\label{sec:introduction}

Motion planning for contact-rich manipulation is a complex task, especially in the presence of non-convexity and hybridness. This requires avoiding poor local optima and optimization over both continuous and discrete variables. We refer to this as contact optimization. In this work, we focus on long-term non-prehensile planar manipulation, which concerns joint logistic and geometric planning and feedback control over diverse interaction modes and face switches. For example, to push an object, a prerequisite is to decide how much force should be applied, and which point to push. Moreover, in some cases such as pushing an object with small distance but large orientation, relying on a single fixed face is not feasible. Therefore, a sequence of face switching is required, as well as contact mode schedule resulting from Coulomb friction.

To solve non-convex problems, trajectory optimization methods are widely used. They iteratively optimize around the current solution till reaching local optima. Although these approaches can yield good practical results, the convergence and solution quality are quite sensitive to the initial guess. The methods can easily get stuck in poor local optima if the initial guess is far from the optimal solution. Moreover, these methods struggle with the hybrid nature of contact. Gradients do not exist when making and breaking contacts, making powerful continuous solvers unusable. One common strategy is to model the contact with complementary constraints on continuous variables \cite{posa2014direct}. However, these constraints can render the optimization problem nearly ill-conditioned, and still susceptible to poor local optima. To address this issue, researchers have proposed warm-starting trajectory optimization methods that provide good initializations to solvers \cite{mansard2018using}. In this paper, we introduce human demonstration as trajectory constraints to the optimization framework, combining the benefits of both approaches. Our approach effectively handles the hybrid variables that arise in contact-rich manipulation tasks while avoiding poor local optima.

\begin{figure} [t]
	\begin{minipage}[t]{.29\linewidth}
	\centering
	\includegraphics[width=1\columnwidth]{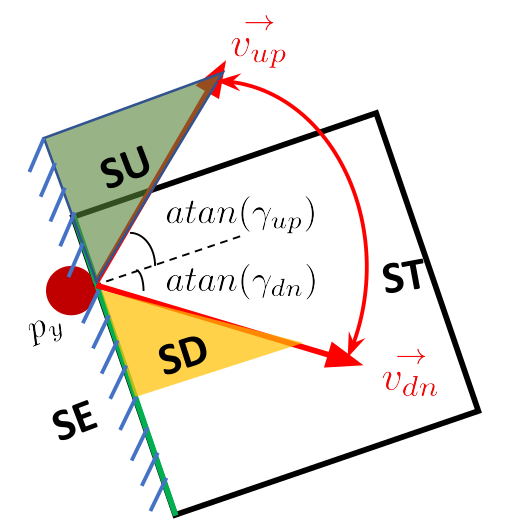}
\end{minipage} 
\hspace{0.1cm}
\begin{minipage}[t]{.31\linewidth}
	\centering
	\includegraphics[width=1\columnwidth]{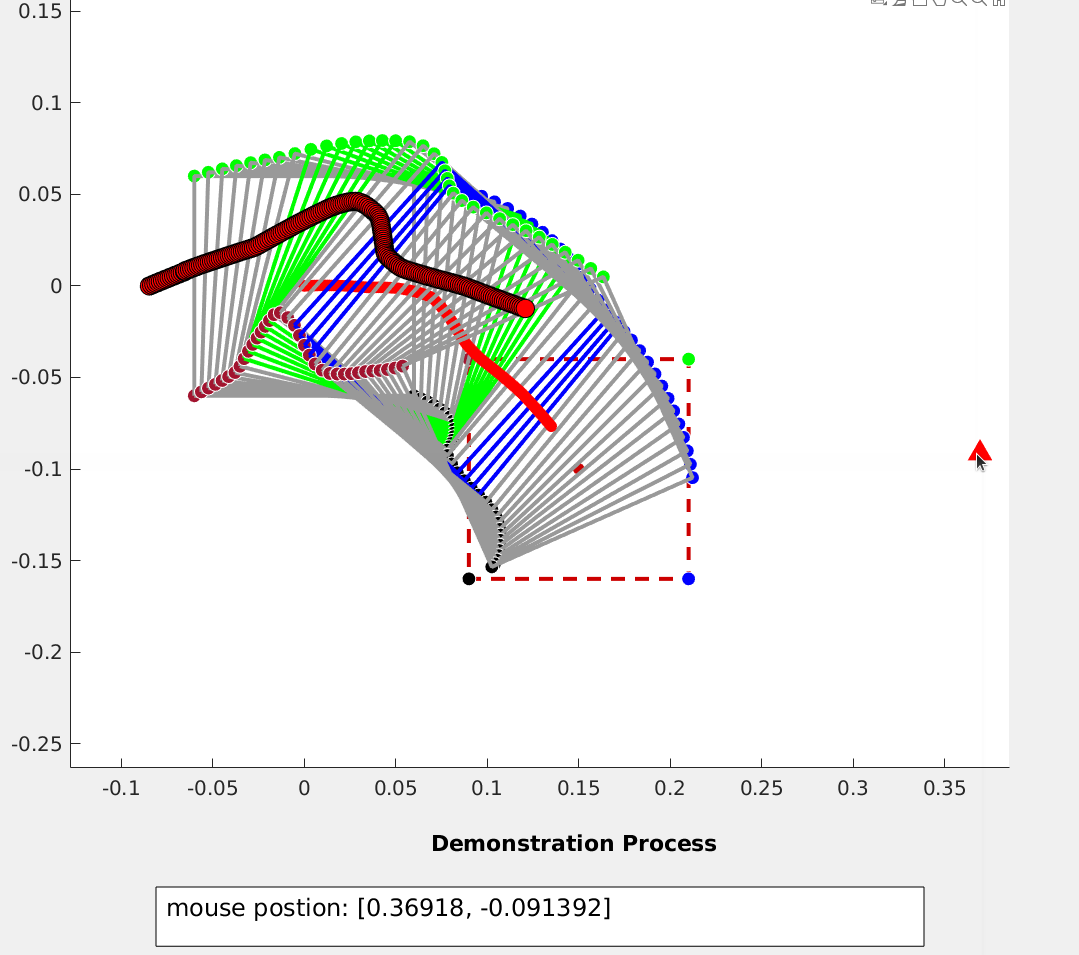}
\end{minipage} 
\hspace{0.1cm}
\begin{minipage}[t]{.31\linewidth}
	\centering
	\includegraphics[width=1\columnwidth]{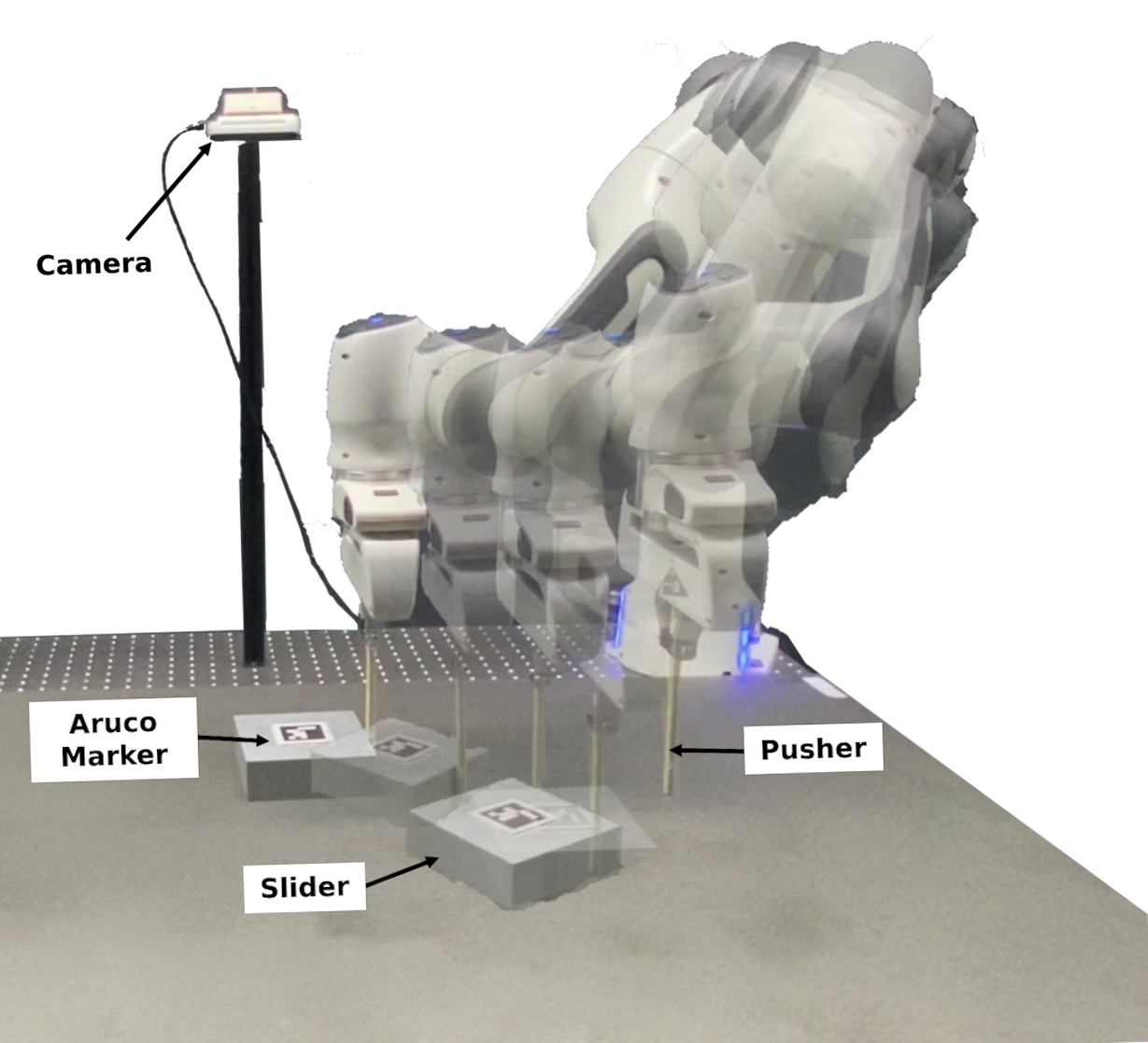}
\end{minipage} 
\hspace{0.1cm}


\caption{\textit{Left}: Multiple contact modes, including sticking (ST), Sliding Up (SU), and Sliding Down (SD) modes for each contact face, as well as a separation (SE) mode for face switching. \textit{Middle}: Demonstration interface. The dotted square represents the target and the red triangle indicates the real-time position of the mouse. Humans can manipulate the mouse to guide the pusher during the demonstration. \textit{Right}: System setup where the robot pushes an object with changes of contact points and faces. } 
\label{fig:pics}
\vspace{-0.3cm}
\end{figure}

\section{Dynamical System}
\label{sec:method1}

We first build the dynamics model for the pusher-slider system with face-switching strategy, namely
\begin{equation}
	\bm{\dot{x}}=\left\{
	\begin{aligned}
		& \bm{g}_1(\bm{x}, \bm{u}), \quad \text{if} \quad \textbf{Sticking}, \\
		& \bm{g}_2(\bm{x}, \bm{u}), \quad \text{if} \quad \textbf{Sliding Up}, \\
		& \bm{g}_3(\bm{x}, \bm{u}), \quad \text{if} \quad \textbf{Sliding Down}, \\
		& \bm{g}_4(\bm{x}, \bm{u}), \quad \text{if} \quad \textbf{Separation},
	\end{aligned}
	\right.
	\label{eq:motion_eq}		
\end{equation}

\noindent where $\bm{x} = [\bm{q}_s^\trsp \ \bm{q}_{p_s}^\trsp \ \bm{v}^\trsp]^\trsp$, representing the slider pose, pusher position and velocity. $\bm{u}$ denotes the acceleration of the pusher. For more information, please refer to \cite{xue2022guided}.

The system, as illustrated in Fig. \ref{fig:pics}-\textit{Left}, consists of three different contact modes (sticking, sliding up and down) for each contact face, along with one separation mode for face switching. Consequently, the problem involves 13 interaction modes and several continuous variables in both state and action spaces.

\section{Methods and Results}
\label{sec:method_results}

\subsection{Problem formulation}
\label{sec:ocp_formu}

An optimal control problem (OCP) can be described as
\vspace{-0.3cm}
\begin{align} \label{eq:basic-cost}
	&\min_{\bm{u}_t} \quad c_T(\bm{x}_T) + \sum_{t=0}^{T-1} c_t(\bm{x}_t,\bm{u}_t) ,\\
	&\begin{array}{r@{\quad}r@{}l@{\quad}l}
		\text{s.t.}  \quad \bm{x}_{t+1} = \bm{f}(\bm{x}_t, \bm{u}_t),
		\label{eq:dynamics}
	\end{array}
\end{align}
\noindent where~\eqref{eq:basic-cost} is the cost function and~\eqref{eq:dynamics} is the dynamic equation. We employ DDP \cite{mayne1966second} to solve this problem, as it can provide a local feedback mechanism for the controller.

\subsection{Demonstration-started DDP (DS-DDP)}
\label{sec:DS_DDP}

Since \eqref{eq:basic-cost} is a non-convex problem, DDP optimizes around the current solution iteratively to solve it. However, as Table \ref{tab:contact_planning} shows, convergence is highly dependent on the initial guess, where \textit{ZS-DDP} represents the typical DDP without demonstration. To address this issue, we incorporate human demonstration as an initialization into the optimization framework. Fig. \ref{fig:pics}-\textit{Middle} shows the demonstration interface, which allows humans to guide the motion of the pusher in real-time using a mouse. The motion of the slider is simulated based on the dynamics model \eqref{eq:motion_eq}. For more details, please see our webpage\footnote{\url{https://sites.google.com/view/dg-oc/}}. We collected demonstration trajectories for both the pusher and slider for three different targets, corresponding to $N_s=0$, $N_s=1$, and $N_s=2$, respectively, where $N_s$ is the number of face switches during the demonstration. Given a target, we use K-nearest neighbor (k-NN) to select the closest demonstration from the set.

The cost function of DS-DDP is defined as
\begin{equation} 
	\begin{gathered}
		DS_{cost} = c_{re} + c_{rg} + c_{bd},
	\end{gathered}
	\label{eq:cost_DS-DDP}
\end{equation}

\noindent where $c_{re} = (\bm{\mu}_T - \bm{x}_T)^\trsp \bm{Q}_T (\bm{\mu}_T - \bm{x}_T)$ represents the reaching cost, $c_{rg} = \sum_{t=0}^{T-1} \bm{u}_t^\trsp \bm{R} \bm{u}_t$ is the regularizer for control commands, and $c_{bd} = \sum_{t=0}^{T-1} \bm{f}^\text{cut}(\bm{u}_t, \bm{u}_l)^\trsp \bm{Q}_f \bm{f}^\text{cut}(\bm{u}_t, \bm{u}_l)$ penalizes control commands through a soft-thresholding function $\bm{f}^\text{cut}$. Here, $\bm{\mu}_T$ is the target state, and $\bm{u}_l$ is the boundary of $\bm{u}$. The initial guess ${\bm{u}^0} = \bm{\widetilde{u}}$ is drawn from human demonstrations directly.

\subsection{Demonstration-penalized DDP (DP-DDP)}
\label{sec:ddp_dem}

The quality and quantity of demonstrations can significantly impact the optimization outcome when using them as the sole initialization. However, demonstrations can also serve as constraints or guidance for optimization, akin to mathematical programming with complementary constraints. Instead of adding slack variables to hard constraints for contact relaxation, we incorporate soft constraints by penalizing the cost function. By leveraging demonstrations as constraints, we can guide the trajectory optimization and mitigate the possibility of getting trapped in poor local optima while effectively handling hybrid variables. It is achieved by designing the cost function as

\begin{equation} 
	DP_{cost} =  c_{re} + c_{rg} + c_{bd} + c_{sw} + c_{ve} + c_{ac},
	\label{eq:cost_demGuided}
\end{equation}
with $c_{sw} = \sum_{n=t_0}^{t_{N-1}}(\bm{\mu}_n - \bm{x}_n)^\trsp \bm{Q}_n (\bm{\mu}_n - \bm{x}_n)$, $c_{ve} = \sum_{t=0}^{T-1} (\bm{\widetilde{v}}_t - \bm{v}_t)^\trsp \bm{R}_{dv} (\bm{\widetilde{v}}_t - \bm{v}_t)$ and $c_{ac} = \sum_{t=0}^{T-1} (\bm{\widetilde{u}}_t - \bm{u}_t)^\trsp \bm{R}_{du} (\bm{\widetilde{u}}_t - \bm{u}_t)$, where $c_{sw}$, $c_{ve}$ and $c_{ac}$ are designed to follow the demonstrated face-switching strategy, pusher velocity and pusher acceleration. $n=[t_0, \cdots, t_{N-1}]$ is the timestep when the contact face switches, $\bm{\mu}$ is the state of the selected demonstration, and $\bm{\widetilde{v}}_t$, $\bm{\widetilde{u}}_t$ are the demonstrated velocity and acceleration at timestep $t$.

\subsection{Warm-starting DDP (WS-DDP)}
\label{sec:warm-start DDP}

Demonstration-penalized DDP shows good performance and can avoid poor local optima, but in order to further improve its convergence properties, we propose a hierarchical optimization framework, where the solution of \textit{DP-DDP} is used to initialize another DDP problem that we call Warm-starting DDP (\textit{WS-DDP}).

The cost function of \textit{WS-DDP} is as same as \textit{DS-DDP}, allowing it to explore much freely towards the final target. The initial guess ${\bm{u}^0} = \bm{u}^*_{DP}$ is the solution of \textit{DP-DDP}.

\subsection{Results}

\renewcommand{\arraystretch}{1.}
\begin{table}[t]
	\caption{Performance of ZS-DDP, DS-DDP, DP-DDP, and WS-DDP for offline programming}
	\begin{footnotesize}
		\begin{tabular}{l | c c c c}
			\toprule
			{\bf{Method}}  & $\bm{x_\text{err}}$ (cm) & $\bm{y_\text{err}}$ (cm) & $\bm{\theta_\text{err}}$ (rad) & $\text{succ\_rate}$ \\
			\midrule
			{ZS-DDP} & 3.18  $\pm$ 8.30  &  4.78  $\pm$  8.60 & 0.04  $\pm$ 0.39 & 40\%\\
			{DS-DDP} & 1.70  $\pm$ 9.07  &  1.62  $\pm$  9.66 & 0.21  $\pm$ 1.82 & 58\%\\
			{DP-DDP} & 0.14  $\pm$ 1.54  &  0.57  $\pm$  2.77 & 0.22  $\pm$ 0.12 & 77\%\\
			{\textbf{WS-DDP}} & \textbf{0.07  $\pm$ 1.31}  &  \textbf{0.19  $\pm$  2.90} & \textbf{0.01  $\pm$ 0.08} & \textbf{85\%} \\
			\bottomrule
		\end{tabular}
	\end{footnotesize}
	\label{tab:contact_planning}
	\vspace{-0.2cm}
\end{table}

To evaluate the effectiveness of our proposed method, we conducted experiments on 100 randomly selected targets from the task space $\mathbb{T} = \{[x, y, \theta]: x \in [\minus 25\text{cm}, 25\text{cm}], y \in [\minus 25\text{cm}, 25\text{cm}], \theta \in [\minus \pi, \pi]\}$. A successful contact planning is achieved when the end-effector reaches the target with an error less than $\{ x_\text{err} < 1\text{cm}, \; y_\text{err} < 1\text{cm}, \; \theta_\text{err} < 5^{\circ} \}$. The statistical results are presented in Table \ref{tab:contact_planning}.

The success rate of our baseline method \textit{ZS-DDP} is only 40\%, indicating that it is susceptible to getting stuck in poor local minima. However, by initializing the optimization with demonstrations, we observe a success rate improvement of nearly 20\%, highlighting the effectiveness of using human demonstrations to avoid poor local minima.

To further enhance the convergence properties of our method, we propose a hierarchical optimization framework consisting of \textit{DP-DDP} and \textit{WS-DDP}. It achieves a significantly higher success rate of 85\% by incorporating human demonstrations as soft constraints and warm-starting a new DDP. By using demonstration as soft constraints, rather than initialization, our approach is less dependent on demonstrations while retaining the prior information. This results in more freedom to optimize around the demonstration trajectory. Our results also show the strong generalization ability of our approach, which only requires a few demonstrations.

The use of human demonstrations in contact-rich manipulation tasks enables traditional trajectory optimization methods to handle both non-convexity and hybridness, opening up promising avenues for further research. For instance, life-long learning from demonstration could be used to learn the constraints from a vast range of daily tasks. These constraints could then be integrated into the optimization framework, enabling the system to develop a dexterous manipulation policy that can handle various tasks robustly.

For a more in-depth understanding of this work, we encourage readers to refer to \cite{xue2022guided}, which presents an extended version with more detailed information.

\newpage

%
\bibliographystyle{IEEEtran}
\bibliography{main}
%

\end{document}